%% file: main36.tex
\documentclass[runningheads]{llncs}
\usepackage{subfig}
\usepackage{makeidx}
\usepackage{graphicx}
\usepackage{amsmath,amssymb} 
\usepackage{color}
\usepackage{pgfplots}
\usepackage{colortbl}
\usepackage{bm}

\usepackage[font=scriptsize,labelfont=bf]{caption}

\usepackage{xspace}

\usepackage{booktabs}	
\usepackage{multirow}

\xdefinecolor{tucol1}{rgb}{1.0 0.15 0.15}
\xdefinecolor{tucol2}{rgb}{0.52,0.72,0.10}
\xdefinecolor{tucol3}{rgb}{0.0 0.0 0.8}
\xdefinecolor{tucol4}{rgb}{0.8 0.8 0}
\xdefinecolor{tucol5}{rgb}{0.8 0 0.8}
\xdefinecolor{tucol6}{rgb}{0 0.8 0.8}

\usepackage{todonotes}

\begin{document}
	\pagestyle{headings}
	\mainmatter

	\def\GCPR17SubNumber{36}

	\title{Neuron Pruning for Compressing Deep Networks using Maxout Architectures}

	\titlerunning{Neuron Pruning for Compressing Deep Networks using Maxout Architectures}
	\authorrunning{Fernando Moya Rueda, Rene Grzeszick, Gernot A. Fink}
	
	\author{Fernando Moya Rueda, Rene Grzeszick, Gernot A. Fink}
	\institute{TU Dortmund University \\ Department of Computer Science}
	

	\maketitle

\begin{abstract}

    This paper presents an efficient and robust approach for reducing the size of deep neural networks by pruning entire neurons. It exploits maxout units for combining neurons into more complex convex functions and it makes use of a local relevance measurement that ranks neurons according to their activation  on the training set for pruning them. Additionally, a parameter reduction comparison between neuron and weight pruning is shown. It will be empirically shown that the proposed neuron pruning reduces the number of parameters dramatically. The evaluation is performed on two tasks, the MNIST handwritten digit recognition and the LFW face verification, using a LeNet-5 and a VGG16 network architecture. The network size is reduced by up to $74\%$ and $61\%$, respectively, without affecting the network's performance. The main advantage of neuron pruning is its direct influence on the size of the network architecture. Furthermore, it will be shown that neuron pruning can be combined with subsequent weight pruning, reducing the size of the LeNet-5 and VGG16 up to $92\%$ and $80\%$ respectively.
    
\end{abstract}

\input{intro}

\input{method}

\input{evaluation}
\input{conclusion}
\clearpage
\bibliographystyle{splncs03}
\bibliography{literatur}

\end{document}

%% file: intro.tex
\section{Introduction}

Having today available a big number of large-scale datasets and powerful GPUs, deep neural networks have become the state-of-the-art in many computer vision, and speech recognition tasks \cite{alvarez2016-LNN,GrzeszickSF16-OPN,Krizhevsky2012-ICD}. 
They achieve high performance in many applications, e.g., scene and object recognition, object detection, scene parsing, face recognition, and medical imaging. 
However, they utilize high computational resources coming along with high memory cost \cite{liu2015-SNN}. For example, AlexNet and DeepFace have around 60M  and 120M parameters, respectively \cite{HanMD15-DCN}. 
Furthermore, they consume significant energy making their application on embedded devices difficult \cite{HanMD15-DCN}. 
Containing a huge amount of parameters, deep neural networks may also be subject to over-parametrization.
Thus, there could exist redundancies, and their generalization is not proper \cite{lecun1989-OBD}. 
In general, networks with a small number of parameters generalize better extracting the important information of the data, rather than over-parametrized networks. 
Nevertheless, smaller networks are harder to train, since they are sensible to initialization \cite{liu2015-SNN}.
    
Designing a network, i.e., setting the number of layers, neurons per layer, and parameters is typically still a "trial and error" process.
Mostly, it depends on experience \cite{alvarez2016-LNN}. 
Moreover, training does not affect the structure of the network \cite{HanPTD15-LWC}. 
Several attempts have been developed for reducing the effect of the huge number of parameters, e.g., dropout \cite{Krizhevsky2012-ICD}, creating an optimal-sized network by adding additional regularizers, or pruning the network parameters \cite{alvarez2016-LNN,lecun1989-OBD,Simonyan2014-VDC}. 
The latter ones attempt to either remove edges or complete neurons from a network. However, most of these approaches require an expensive comparison of all neurons in the network or additional and expensive post-processing, e.g., the computation of the network's Hessian.
    
We propose an efficient and robust method for neuron pruning based on a local decision for reducing the number of parameters in a deep neural network. 
We will empirically show that pruning neurons rather than weights is essential for reducing the size of a neural network at runtime.
The proposed approach for pruning neurons is based on the good performance of maxout units \cite{Goodfellow2013-MN}, which were developed for boosting the impact of dropout in training, and on their capacity to combine neurons for approximating more complex functions.
It is assumed that redundancies typically exist in a neural network, so they also exist in a maxout unit, which combines the output of multiple neurons.
Thus, pruning can be performed in a very local approach based on a single maxout unit.

The remainder of the paper is structured as follows:
section \ref{sec:related} will discuss the related work in the field of parameter pruning for deep networks.
In section \ref{sec:review} and \ref{sec:method}, the maxout approach will be reviewed and an approach for pruning neurons will be introduced.
Experiments on two datasets, the MNIST digits dataset and the labeled faces in the wild dataset, will be shown in section \ref{sec:eval}.
Face recognition has been chosen as an application that is of special interest for embedded devices such as smartphones.
The last section presents a short conclusion.

\section{Related Work}
\label{sec:related}

Though deep neural networks are very powerful, they are known to be over-parametrized, possessing millions of parameters \cite{liu2015-SNN}.
This over-parametrization may cause performance deficits, e.g., poor generalization, overfitting, slow testing time, and enormous energy and memory consumption \cite{denil2013-PPL,HanMD15-DCN,HanPTD15-LWC,lecun1989-OBD}.
Therefore, reducing the size of the network by removing unimportant parameters, or designing optimal-sized networks becomes imperative. 
For those purposes, different attempts have been developed. 
These attempts can be grouped in constructive and destructive methods.
    
In constructive methods, neurons or layers are added to a trained shallow neural network. 
For example, in \cite{Simonyan2014-VDC} a very deep convolutional neural network (CNN) is trained by continuously adding convolutional layers to an initial CNN of $11$ layers for obtaining a better performance. 
However, the initial shallow networks must be properly trained, as the network can otherwise get stuck into a local optimum. 
Moreover, as the idea is to improve the network's performance by adding layers and neurons, the network's size increases, and redundancies could be introduced into the network.
    
In destructive methods, non-relevant neurons (neuron pruning) and/or parameters (weights pruning) of an initial deep neural network are removed, while maintaining its behaviour. 
The authors in \cite{lecun1989-OBD} and \cite{mozer1989-SNN} started the concept of pruning neural networks, both using a sort of relevance measure. In \cite{lecun1989-OBD}, parameters, with the smallest relevance in the network, which is computed by using the Hessian of the loss function, are deleted. 
In \cite{mozer1989-SNN}, complete neurons are deleted by using a relevance measurement based on the difference of the network's performance with and without the neuron. 
Nevertheless, especially with today's very deep neural networks with millions of parameters, computing the relevance of each neuron or parameter demands very high computational resources.
    
Different from the relevance measure methods, the authors in \cite{HanMD15-DCN,HanPTD15-LWC} prune weights by thresholding them.
Afterwards, the network is re-trained for compensating the lost connections. One of the most prominent destructive methods is Deep Compression \cite{HanMD15-DCN}. 
The authors reduced the storage required for a deep CNN by a factor of $35$ and $49$. 
They used a combination of three steps: weight pruning, weight quantization and Huffman coding.
First, weight pruning is applied by thresholding the weights and thus setting them to zero.
The remaining weights are then quantized, which reduces the number of bits for representing weights.
Finally, a Huffman coding is applied.
However, the networks are currently de-compressed for inference.
    
Recently, the authors in \cite{alvarez2016-LNN} determined the best number of parameters, by using a regularizer while training the network. 
The regularizer forces all the weights of single neurons to be zero. 
For testing, these dead neurons are removed from the network. 
Nevertheless, additional hyper-parameters must be determined for the regularization. 
The authors in \cite{denil2013-PPL} reduced the number of parameters to be learned by factorizing the weight matrices as a low rank product of two matrices: a static, and a dynamic matrix. 
First, they trained the static matrices as a general dictionary, obtaining a prior knowledge of the smoothness structures that are expected to be seen. Second, they fine-tuned the dynamic matrix, which are the weights to be learned. 
        
\begin{figure}[t!]
	\centering
    \subfloat[LeNet-5]{ \includegraphics[height=13em]{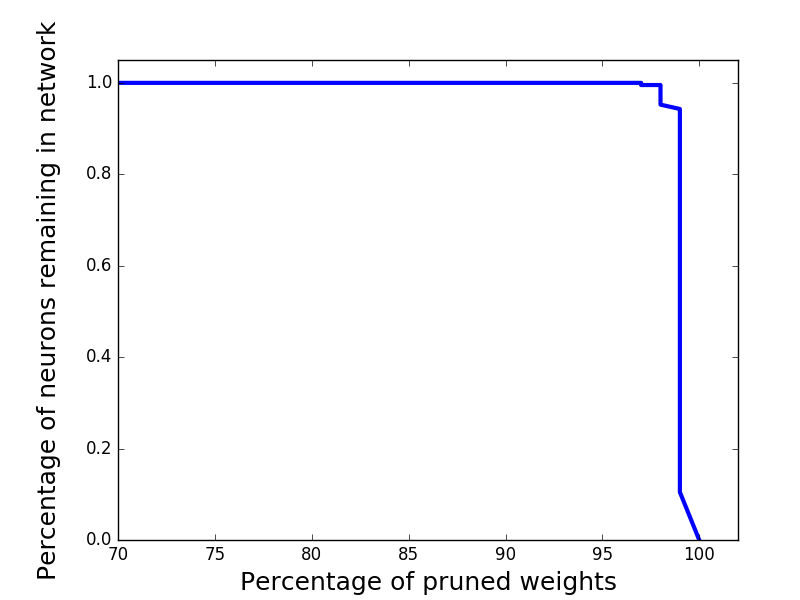} }
    \hspace*{-2em}
    \subfloat[VGG16]{ \includegraphics[height=13em]{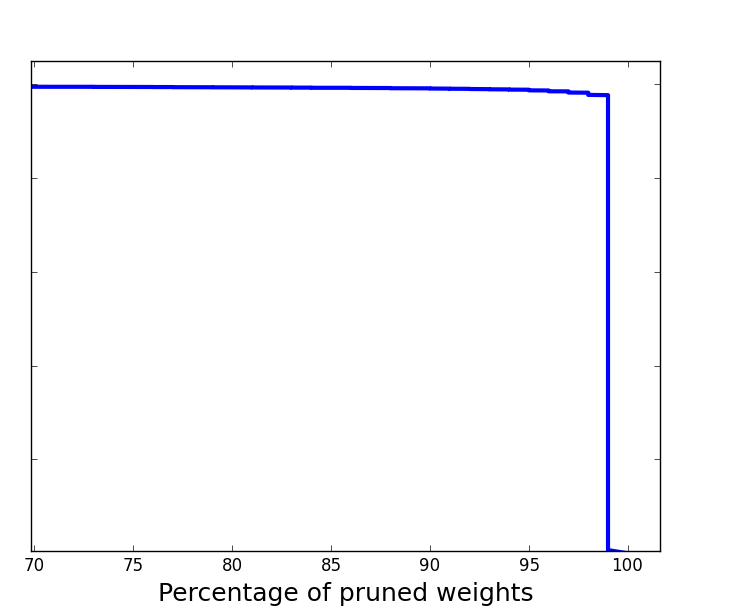} }
    \caption{Percentage neurons  remaining in network vs percentage of pruned weights for a LeNet-5 trained for digit recognition and for a VGG16 trained for face recognition (for details see section \ref{sec:eval}).}
    \label{fig:weightPruning_intro}%
\end{figure}

Comparing the approaches, the destructive methods are more popular than the constructive ones as the networks are often easier to train.
Good compression results are achieved by the deep compression approach in \cite{HanMD15-DCN}.
However, pruning weights has the disadvantage of rarely removing neurons from the network architecture.
To make this clear, we show in Fig.~\ref{fig:weightPruning_intro} the relation between the proportion of remaining neurons in the network versus the proportion of pruned weights using the LeNet-5 \cite{lecun1998-GBL} and the VGG16 \cite{Parkhi15-DFR}  as examples.
It clearly shows that neurons only get pruned at a very high compression ratio, which typically influences the networks performance.
Thus, thresholding parameters allows for compressing a network but does not influence the size of the architecture.
At runtime a sparse matrix library would be required, which efficiently evaluates the compressed network.
Otherwise, the network must be de-compressed for inference. The zeroed weights are again stored in the memory, as in \cite{HanMD15-DCN}, which generates a waste in memory consumption as well as computational power.

%% file: method.tex
\section{Fundamentals}
\label{sec:review}

The proposed approach builds on a maxout architecture \cite{Goodfellow2013-MN} for pruning the neural networks in a destructive manner. 
A maxout layer can be considered as a cross-channel pooling operation, performing a max operation between $k$ adjacent neurons. 
These layers were designed to boost the model's averaging ability of dropout \cite{Krizhevsky2012-ICD}, thought for preventing overfitting, and to improve the optimization. 
Given an input layer $X=[x_0,x_1,x_2,..., x_N]$ with $N$ neurons, a maxout layer computes:
\begin{equation}
	h(X) = \max [x_{jk+0}, x_{jk+1}, x_{jk+2},..., x_{jk+(k-1)}] \quad \forall j\in [0,N/k-1],
\end{equation}
where $k$ is the number of neurons that are combined into a single maxout unit. 

As the authors in \cite{Goodfellow2013-MN} show, a maxout unit is a universal approximator. It combines $k$ single neurons implementing a piecewiese linear function that can approximate arbitrary convex functions. So, in theory, the maximum of several neurons is able to approximate a more complex neuron. Moreover, the maxout unit becomes a sort of an activation function, replacing other activation functions, but with a factor of $k$ smaller number of parameters. For example, two linear functions can implement a ReLU function, or five different linear functions can implement an approximation of a quadratic one, as shown in \cite{Goodfellow2013-MN}. 

    
\section{Compressing Networks with Maxout Architectures}
\label{sec:method}
The idea of the proposed approach is to use the maxout units and their model selection abilities for pruning entire neurons from an architecture without expensive processing.
Thus, reducing the size and the memory consumption of a deep network. 
In some cases, the performance of the network may even increase as redundancies get reduced or eliminated.
    
Following the assumption that redundancies exist in a deep neural network, it is assumed that if a network contains a maxout layer, redundancies will, also, exist in the maxout units.
This is a valid assumption, since dropout and other regularization approaches cause the learn process to create different paths through the deep network, which yield similar outputs \cite{Goodfellow2013-MN}.
Using this premise, a reduction of the number of neurons in a maxout layer can be done without an expensive relevance measurement.

\subsection{Neuron Pruning} \label{neuronpruning}

\begin{figure}[t]
            \centering
                \begin{tikzpicture}[thick,scale=0.9, every node/.style={scale=0.8}]
                \tikzstyle{arrow} = [->, line width=1, thick]
                \tikzstyle{circle1}=[circle,draw=black, thick]
                \tikzstyle{rectangle1}=[rectangle,draw=black]
                \tikzstyle{node1}=[text=black, font=\footnotesize \bfseries];
                \tikzstyle{node2}=[text=black, font=\footnotesize \bfseries];
                \tikzstyle{node3}=[text=red, font=\scriptsize \bfseries];
                \tikzstyle{node4}=[text=black, font=\tiny \bfseries];
                \tikzstyle{node5}=[text=black, font=\tiny \bfseries];

                \node [node1](v1) at (-5.0,1.8) {$x_0^0$};
                \node [node1](v3) at (-5.0,1.2) {$x_1^0$};
                \node [node1](v4) at (-5.0,0.6) {$x_2^0$};
                \node [node1](v5) at (-5.0,0.1) {$x_3^0$};
                \node [node3](v6) at (-5.0,2.2) {$\textrm{No pruning}$};
                \node (v2) at (-4,1) {};
                \node [circle1] (c1) at (-4,1) {};
                \node [node5] at (-3.7,1.5) {Max unit};
                \draw  [arrow](v1) edge (v2);
                \draw  [arrow](v3) edge (v2);
                \draw  [arrow](v4) edge (v2);
                \draw  [arrow](v5) edge (v2);
                
                \node [rectangle1] (r1) at (-2.2,0.6) [minimum width=2cm,minimum height=1.2cm]{};
                \node [node4] at (-2.3, 1.0) {Remove least};
                \node [node4] at (-2.25, 0.8) {active neuron};
                \node [node4] at (-2.6, 0.5) {i.e.,$x_2^0$};
                \node [node4] at (-2.5, 0.2) {$\rightarrow$ re-train};

                \node [node1](v10) at (-0.5,1.8) {$x_0^1$};
                \node [node1](v11) at (-0.5,1.2) {$x_1^1$};
                \node [node1](v13) at (-0.5,0.1) {$x_3^1$};
                \node [node3](v15) at (-0.5,2.2) {$\textrm{Iteration 1}$};
                \node (v14) at (0.5,1) {};
                \node [circle1] (c10) at (0.5,1) {};
                \node [node5] at (0.8,1.5) {Max unit};
                \draw  [arrow](v10) edge (v14);
                \draw  [arrow](v11) edge (v14);
                \draw  [arrow](v13) edge (v14);
                
                \node [rectangle1] (r2) at (2.0,0.6) [minimum width=2cm,minimum height=1.2cm]{};
                \node [node4] at (1.9, 1.0) {Remove least};
                \node [node4] at (1.95, 0.8) {active neuron};
                \node [node4] at (1.6, 0.5) {i.e.,$x_1^1$};
                \node [node4] at (1.7, 0.2) {$\rightarrow$ re-train};
                
                \node [node1](v21) at (3.5,1.8) {$x_1^2$};
                \node [node1](v23) at (3.5,0.1) {$x_3^2$};
                \node [node3](v25) at (3.5,2.2) {$\textrm{Iteration 2}$};
                \node (v24) at (4.5,1) {};
                \node [circle1] (c20) at (4.5,1) {};
                \node [node5] at (5.0,1.5) {Max unit};
                \draw  [arrow](v21) edge (v24);
                \draw  [arrow](v23) edge (v24);
                
                \node [rectangle1] (r2) at (6.0,0.6) [minimum width=2cm,minimum height=1.2cm]{};
                \node [node4] at (5.9, 1.0) {Remove least};
                \node [node4] at (5.95, 0.8) {active neuron};
                \node [node4] at (5.6, 0.5) {i.e.,$x_3^2$};
                \node [node4] at (5.7, 0.2) {$\rightarrow$ re-train};
                
                \node [node1](v31) at (7.5,1.0) {$x_1^3$};
                \node [node3](v35) at (7.5,2.2) {$\textrm{Iteration 4}$};  
                \node (v34) at (8.5,1) {};
                \draw  [arrow](v31) edge (v34);
                
                \end{tikzpicture}
            \caption{Neuron pruning process for $k=4$ inputs per maxout unit. $x_a^b$ represents a neuron with $'a'$ the neuron index and $'b'$ the iteration.}
            \label{fig:neuronPruningProcess}
        \end{figure}
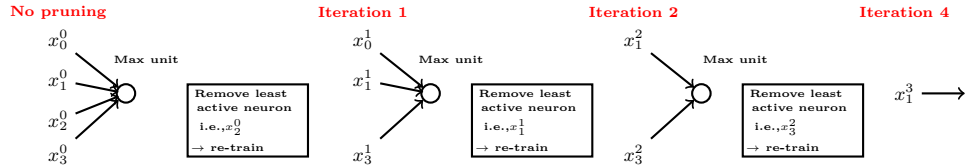

For reducing the size of a CNN using maxout units, an iterative process is followed. 
First, a CNN with a maxout layer is trained. 
This maxout layer performs a max function among $k$ adjacent neurons, reducing the amount of weights connecting with the next layer by a factor of $k$. So, placing this maxout layer after the one with the highest number of weights would be advisable.
Second, by counting the number of times neurons become the maximal value in each maxout unit when computing a forward pass over the training dataset, the least active neurons of each maxout unit are removed from the network.
Their effects are negligible with respect to other neurons.
Third, the remaining neurons of the CNN are re-trained.
After re-training, the process is repeated; in this case, the maxout layer performs a max function among $k-1$ neurons, and so on.
Fig.~\ref{fig:neuronPruningProcess} shows an example for $k=4$.
        
In comparison with \cite{mozer1989-SNN}, pruning neurons takes place locally, since relevance values are not computed depending on the network's output for each single neuron. 
The pruning in maxout architectures is therefore more feasible for very large networks with millions of parameters.

\subsection{Weight Pruning} 
\label{weightpruning}
Having reduced the number of parameters in the network by pruning neurons from the maxout units of the network, further compression operations can be performed.
Following the approach in \cite{HanMD15-DCN,HanPTD15-LWC}, connections (weights) can be pruned in an additional processing step.
Based on thresholding, edges with lower value than a threshold are set to zero. 
Thus, learning which connections are important and deleting the unimportant ones. 
By this weight pruning, the network becomes a sparse network. 
For pruning weights, a three-step procedure is followed, as proposed in \cite{HanMD15-DCN,HanPTD15-LWC}. 
Given the network that has been compressed by the proposed neuron pruning, the important connections are learned based on a global threshold.
The threshold can be set such that as many connections as possible are removed without deteriorating the performance on a validation set.
Second, weights below this threshold are deleted; that is, weights are set to zero. 
Third, the network is re-trained, learning the final weights.

%% file: evaluation.tex
\section{Evaluation}
\label{sec:eval}
An evaluation of both neuron and weight pruning is carried out for two different tasks: handwritten digit recognition, MNIST dataset \cite{lecun1998-GBL}, and face verification, LFW dataset \cite{Huang2007-LFW}. 
The later is of special interest for embedded domains, e.g., in mobile phones. 
In general, the performance of the networks is evaluated with a varying percentage of pruned weights: after applying maxout, when pruning several neurons from the maxout units, and finally after applying additional weight pruning.
While in the first task a very small LeNet-5 architecture is compressed, in the second task a large VGG16 architecture is compressed.

For the experiments, we chose $k=4$ for the size of the maxout units as it allows for a fairly good compression and does not reduce the descriptiveness of the network compared to a network without maxout units.
The neurons are then iteratively pruned from the maxout units and the network is re-trained after each pruning step.

\subsection{Handwritten digit recognition} 
For the digit-recognition task, two networks, using the LeNet-5 architecture \cite{lecun1998-GBL} with two convolutional layers, a fully connected layer and a softmax layer as a classificator, were trained. One network contains a maxout layer after the fully connected layer (LeNet-MFC), while the other has a maxout layer after the last convolutional layer (LeNet-MC). An iterative training following the steps in section~\ref{neuronpruning} is executed using the MNIST dataset \cite{lecun1998-GBL}. This dataset consists of $60000$ handwritten-digit images (of size $[28\times28]$) for training and $10000$ images for testing. 
We used stochastic gradient descent (SGD) with a momentum of 0.9, weight decay of $5 \times 10^{-4}$ with inverse decay, a base learning rate of $0.01$ that is iteratively reduced and a batch size of $64$ for training. The networks were trained for $10000$ iterations.
        
Table \ref{tab:maxoutMNIST} shows the classification accuracy for both networks with different fully connected layer sizes, with and without maxout (after the fully connected layer or the last convolutional layer).
Pruning of one up to three neurons is evaluated. It shows also the percentage of pruned weights which do not remain in the network's architecture, denoted by $p.w.\%$. 
In general, for both networks when using maxout and pruning neurons, the accuracy is maintained. The slight deviations of the accuracies of both networks with respect to the original networks are not significant based on a randomization test \cite{Ojala2010-PTF}. Moreover, the number of weights are considerably reduced with up to $70\%$ for LeNet-MFC and $74\%$ for LeNet-MC. However, this reduction changes with respect to the position of the maxout layer. In LeNet-MFC, each neuron pruning step reduces the number of weights by $19.8\%$, because the neurons are pruned from the fully connected layer, which has the largest number of weights in the network. Besides, the maxout layer does not provide a considerable reduction, since it reduces the size of the softmax layer that has less number of weights compared with the other layers. In contrast, the weight reduction in LeNet-MC due to neuron pruning is just $1.4\%$ per step, and it comes mostly from the maxout layer. In this case, the maxout layer reduces the fully connected layer instead, and the neurons are pruned from the last convolutional layer. However, in the last convolutional layer the number of weights is negligible. 

        \begin{table}[t]
            \scriptsize
            \hspace*{-0.7em}
            \centering
            \setlength{\tabcolsep}{2pt}
            \renewcommand{\arraystretch}{1.2}
                \begin{tabular}{c c||c||c c||c c||c c||c c}
                    \multicolumn{2}{c||}{Network} & No Maxout & \multicolumn{2}{c||}{No prun} &\multicolumn{2}{c||}{1 neuron prun.} & \multicolumn{2}{c||}{2 neuron prun.} &\multicolumn{2}{c|}{3 neuron prun.}\\
                    \hline
                    Type & FC size & Acc$\%$ & Acc$\%$ & p.w.$\%$ & Acc$\%$ & p.w.$\%$ & Acc$\%$ & p.w.$\%$ & Acc$\%$ & p.w.$\%$ \\
                    \hline
                    \multirow{3}{*}{LeNet-MFC} & 128 & 98.1 & 99.1 & 0.74 & 99.1 & 22.6 & 99.1 & 40.4 & 99.1 & 60.2\\
                    \cline{2-11}
                          & 256 & 98.3 & 99.2 & 0.82 & 99.2 & 22.8 & 99.2 & 44.8 & 99.0 & 66.8\\
                    \cline{2-11}
                          & 512 & 99.1 & 99.2 & 0.87 & 99.2 & 24.1 & 99.2 & 47.4 & \textbf{99.1} & \textbf{70.7} \\
                    \hline
                    \hline
                    \multirow{3}{*}{LeNet-MC} & 128 & 98.1 & 99.2 & 59.4 & 99.2 & 64.2 & 99.2 & 69.1 & 99.0 & 73.9\\
                    \cline{2-11}
                     & 256 & 98.3 & 99.0 & 65.9 & 99.2 & 68.6 & 99.3 & 71.3 & 99.0 & 73.9\\
                    \cline{2-11}
                          & 512 & 99.1 & 99.2 & 69.7 & 99.3 & 71.1 & \textbf{99.3} & \textbf{72.6} & \textbf{99.3} & \textbf{74.0}\\
                \end{tabular}
            \vspace*{1em}
            \caption{Accuracies in [$\%$] and pruned weight's proportions (p.w.$\%$) in [$\%$] for LeNet-5 with Maxout layer ($k=4$) after last fully connected layer (LeNet-MFC) and last convolutional layer (LeNet-MC).}
            \label{tab:maxoutMNIST}
        \end{table}
            
        \begin{table}[tb]
            \scriptsize
            \centering
            \setlength{\tabcolsep}{4pt}
            \renewcommand{\arraystretch}{1.1}
            \begin{tabular}{c c||c c c c c c c c c}
  
                    \multicolumn{2}{c||}{Network} &\multicolumn{9}{c}{proportions in $[\%]$ of pruned weights}\\
                    Type & FC size & 0 & 10 & 30 & 50 & 60 & 70 & 80 & 90 & 98\\
                    \hline
                    \multirow{3}{*}{LeNet-MFC} & 128 & 99.0 & 99.0 & 99.0 & 99.0 & 99.0 & \textbf{99.0} & 98.9 & 98.3 & 82.1\\
                    \cline{2-11}
                    & 256 & 99.3 & 99.3 & 99.3 & 99.3 & \textbf{99.3} & 99.2 & 99.2 & 98.9 & 91.0\\
                    \cline{2-11}
                    & 512 & 99.3 & 99.3 & 99.3 & 99.3 & 99.3 & \textbf{99.3} & 99.2 & 99.1 & 92.2\\
                    \hline
                    \hline
                    \multirow{3}{*}{LeNet-MC} & 128 & 99.2 & 99.2 & 99.2 & 99.2 & 99.1 & 99.1 & 98.9 & 97.8 & 26.5\\
                    \cline{2-11}
                    & 256 & 99.3 & 99.3 & 99.3 & 99.2 & 99.2 & 99.1 & 98.9 & 96.3 & 55.3\\
                    \cline{2-11}
                    & 512 & 99.2 & 99.2 & 99.2 & 99.2 & 99.2 & \textbf{99.2} & 99.0 & 97.4 & 60.8\\
            \end{tabular}
            \vspace*{1em}
            \caption{Accuracies in [$\%$] for both networks, after pruning three neurons out of four, under different proportions of pruned weights.}
            \label{tab:weightPruning}
        \end{table}

Following the neuron pruning, additional weight pruning, as discussed in section \ref{weightpruning}, can be applied.
        As mentioned in \cite{HanPTD15-LWC}, neurons could also be pruned from the network if all their input weights are zero; that is, the neuron can be considered as \textit{dead}. 
        So, the number of neurons, and thus the number of weights, could be considerably reduced if a proper threshold is used. 
        Nevertheless, analyzing the proportion of dead neurons versus the proportion of pruned weights, neurons do not become \textit{dead} before pruning more than $98\%$ of the weights, see Fig.~\ref{fig:weightPruning_intro}(b). 
        Consequently, weight pruning rarely prunes neurons. 
        Thus, zeroed weights remain in the network as part of the neurons and the network's architecture does not change so that a sparse representation would be required at runtime \cite{HanMD15-DCN}.
However, assuming the usage of such a representation and for reducing storage size of the network, additional weight pruning is applied to both networks.
        As a basis, we use the networks after pruning three out of four neurons in the maxout units.
        The results in  Tab.~\ref{tab:weightPruning} show that the accuracy will not drop if less than $70\%$ of the weights are thresholded. So, a total compression rate of \textbf{$91\%$} for LeNet-MFC and \textbf{$92\%$} for LeNet-MC, of pruned and zeroed weights, can be reached.

\subsection{Face Verification}
        
\begin{figure}[t]
            \centering
            \begin{multicols}{2} 
          		\begin{tikzpicture}
                \tikzstyle{arrow} = [line width=1, thick]
                \tikzstyle{arrow1} = [->, line width=1, thick]
                \tikzstyle{arrow2} = [line width=1, draw=red, very thick]
                \tikzstyle{circle1}=[circle,draw=black, thick]
                \tikzstyle{node1}=[text=black, font=\footnotesize \bfseries];
                \tikzstyle{node2}=[text=black, font=\tiny \bfseries];
                \tikzstyle{node3}=[text=black, font=\scriptsize \bfseries];
                
                \draw [fill=white] (0.0,0.3) rectangle +(1,1);
                \draw [fill=white] (-0.1,0.2) rectangle +(1,1);
                \draw [fill=white] (-0.2,0.1) rectangle +(1,1);
                \draw [fill=white] (-0.3,0) rectangle +(1,1);
				\node [node1] at (0.2,-0.7){$conv5\_3$};
				\node [node3] at (0.2,-1.0){p.w.$1.62\%$};
				\node [node3] at (-0.4,1.25){$512$};
				\node [node2] at (1.0,-0.1){$\textrm{max}$};
				\node [node2] at (1.0,-0.25){$\textrm{pooling}$};		
				
                \node [node1](v13) at (1,0.65){};
                \node [node1](v14) at (2,0.65){};
                \draw  [arrow1](v13) edge (v14);
                \node [node3] at (1.5,0.45){$dense$};
                
				\draw [fill=white] (2,-0.3) rectangle +(0.4,1.9);
				\node [node1] at (2.2,-0.7){$fc6$};
				\node [node3] at (2.2,-1.0){p.w.$70.1\%$};
				\node [node3] at (2.2,1.75){$4096$};
				
                \node [node1](v1) at (2.7,1.5){};
                \node [node1](v3) at (2.7,1.3){};
                \node [node1](v4) at (2.7,0.9){};
                \node [node1](v5) at (2.7,0.7){};
                \node (v2) at (3.5,1.1) {};
                \node [circle1] (c1) at (3.5,1.1) {};
                \draw  [arrow](v1) edge (v2);
                \draw  [arrow](v3) edge (v2);
                \draw  [arrow](v4) edge (v2);
                \draw  [arrow](v5) edge (v2);
                
                \node [node1](v6) at (2.7,0.6){};
                \node [node1](v8) at (2.7,0.4){};
                \node [node1](v9) at (2.7,0.0){};
                \node [node1](v10) at (2.7,-0.2){};
                \node (v7) at (3.5,0.2) {};
                \node [circle1] (c1) at (3.5,0.2) {};
                \draw  [arrow](v6) edge (v7);
                \draw  [arrow](v8) edge (v7);
                \draw  [arrow](v9) edge (v7);
                \draw  [arrow](v10) edge (v7);
                \node [node2] at (3.5,1.6){maxout units};
                \node [node2] at (3.5,1.8){$4096/k$};
                
                \node [node1](v11) at (3.5,0.65){};
                \node [node1](v12) at (4.5,0.65){};
                \draw  [arrow1](v11) edge (v12);
                \node [node3] at (4.0,0.45){$dense$};
				
				\draw [fill=white] (4.5,-0.3) rectangle +(0.4,1.9);
				\node [node1] at (4.7,-0.7){$fc7$};
				\node [node3] at (4.7,-1.0){p.w.$11.6\%$};
				\node [node3] at (4.7,1.75){$4096$};
				
                \node [node1](v15) at (5.5,-1.2){};
                \node [node1](v16) at (5.5,2.3){};
                \draw  [arrow2](v15) edge (v16);

                \end{tikzpicture}

                \begin{tikzpicture}
                \tikzstyle{arrow} = [line width=1, thick]
                \tikzstyle{arrow1} = [->, line width=1, thick]
                \tikzstyle{arrow2} = [line width=1, draw=red, very thick]
                \tikzstyle{arrow3} = [line width=1, draw=white, very thick]
                \tikzstyle{circle1}=[circle,draw=black, thick]
                \tikzstyle{node1}=[text=black, font=\footnotesize \bfseries];
                \tikzstyle{node2}=[text=black, font=\tiny \bfseries];
                \tikzstyle{node3}=[text=black, font=\scriptsize \bfseries];
                
                \draw [fill=white] (0.0,0.3) rectangle +(1,1);
                \draw [fill=white] (-0.1,0.2) rectangle +(1,1);
                \draw [fill=white] (-0.2,0.1) rectangle +(1,1);
                \draw [fill=white] (-0.3,0) rectangle +(1,1);
				\node [node1] at (0.2,-0.7){$conv5\_2$};
				\node [node3] at (0.2,-1.0){p.w.$1.62\%$};
				\node [node3] at (-0.4,1.25){$512$};	
				
                \node [node1](v13) at (1,0.65){};
                \node [node1](v14) at (2,0.65){};
                \draw  [arrow1](v13) edge (v14);
                \node [node2] at (1.5,0.45){$convs$};
                
                \draw [fill=white] (2.3,0.3) rectangle +(1,1);
                \draw [fill=white] (2.2,0.2) rectangle +(1,1);
                \draw [fill=white] (2.1,0.1) rectangle +(1,1);
                \draw [fill=white] (2.0,0) rectangle +(1,1);
				\node [node1] at (2.7,-0.7){$conv5\_3$};		
				\node [node3] at (2.7,-1.0){p.w.$1.62\%$};
				\node [node3] at (1.9,1.25){$512$};
				\node [node2] at (3.0,-0.1){$\textrm{max}$};
				\node [node2] at (3.0,-0.25){$\textrm{pooling}$};

                \node [node1](v1) at (3.4,1.5){};
                \node [node1](v3) at (3.4,1.3){};
                \node [node1](v4) at (3.4,0.9){};
                \node [node1](v5) at (3.4,0.7){};
                \node (v2) at (4.2,1.1) {};
                \node [circle1] (c1) at (4.2,1.1) {};
                \draw  [arrow](v1) edge (v2);
                \draw  [arrow](v3) edge (v2);
                \draw  [arrow](v4) edge (v2);
                \draw  [arrow](v5) edge (v2);
                
                \node [node1](v6) at (3.4,0.6){};
                \node [node1](v8) at (3.4,0.4){};
                \node [node1](v9) at (3.4,0.0){};
                \node [node1](v10) at (3.4,-0.2){};
                \node (v7) at (4.2,0.2) {};
                \node [circle1] (c1) at (4.2,0.2) {};
                \draw  [arrow](v6) edge (v7);
                \draw  [arrow](v8) edge (v7);
                \draw  [arrow](v9) edge (v7);
                \draw  [arrow](v10) edge (v7);
                \node [node2] at (4.2,1.6){maxout units};
                \node [node2] at (4.2,1.8){$4096/k$};
                
                \node [node1](v11) at (4.2,0.65){};
                \node [node1](v12) at (5.2,0.65){};
                \draw  [arrow1](v11) edge (v12);
                \node [node3] at (4.7,0.45){$dense$};
				
				\draw [fill=white] (5.2,-0.3) rectangle +(0.4,1.9);
				\node [node1] at (5.4,-0.7){$fc6$};
				\node [node3] at (5.4,-1.0){p.w.$70.1\%$};
				\node [node3] at (5.4,1.75){$4096$};
				
                \node [node1](v15) at (6.0,-1.2){};
                \node [node1](v16) at (6.0,2.3){};
                \draw  [arrow3](v15) edge (v16);
                
                \end{tikzpicture}
                
            \end{multicols}
            \caption{Comparison of two architectural approaches for placing the maxout units: (left) VGG16-MFC, (right) VGG16-MC. The weight proportions p.w.$\%$ per layer are also shown.}
            \label{fig:maxoutNetworks}
        \end{figure}
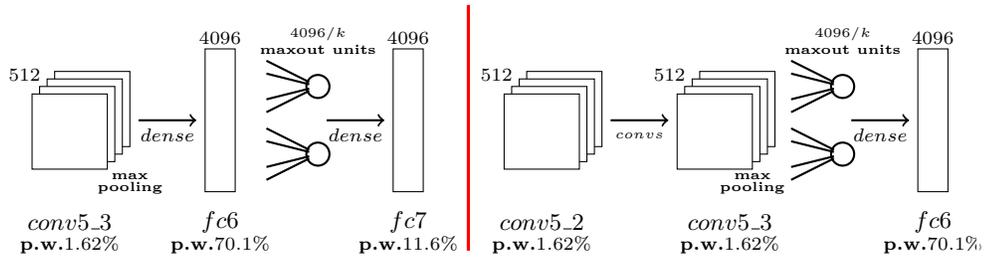

The neuron pruning was also carried out for a larger network for the purpose of face verification. The task is to verify whether two face-images portray the same person or not. For that purpose, the VGG16 network \cite{Parkhi15-DFR} was utilized, using The Visual Geometry Group Face Dataset (VGG face-dataset) as a training-dataset. This dataset is a large collection of face-images containing $2.6$ million face-images from $2622$ identities. It does not contain overlapping identities with standard benchmark datasets (LFW, YFT), so it is suitable for training. The VGG16 network, configuration D in \cite{Parkhi15-DFR}, is a deep CNN with 16 layers: $13$ convolutional layer, two fully connected layers, and a softmax layer. Analogous to the previous LeNet configurations, two configuration of VGG16 are used, in which a maxout network with $k=4$ is added after the first fully connected layer ($fc6$), called VGG16-MFC, and after the last convolutional layer ($conv\_5$), called VGG16-MC, see Fig.~\ref{fig:maxoutNetworks}. The positions of the maxout layers are set after the layers with the most quantity of weights. 
        Since, the connections between $conv\_5$ and $fc6$ have $70.1\%$ of the total amount of weights in the network and the connections between  $fc6$ and $fc7$ have an additional $11.6\%$ of the network's weights. The last three fully connected layers were fined-tuned for both networks. In the case of VGG16-MC, the $conv\_5$ was also fine-tuned. We used SGD with a momentum of $0.9$, weight decay of $5\times 10^{-4}$, three learning rates $[10^{-2}, 10^{-3}, 10^{-4}]$, as \cite{Parkhi15-DFR}, and a batch size of $128$.

        The network was tested following the procedure in \cite{Parkhi15-DFR}, but using the restricted configuration of The Labeled Faces in the Wild (LFW) \cite{Huang2007-LFW}. The LFW dataset is a standard benchmark dataset for face verification. It contains $13233$ face-images from $5749$ identities extracted from the Internet. Faces in images were detected using the Viola-Jones face detector \cite{Huang2007-LFW}. Faces are roughly centered, contain lesser noise but larger bounding-box than the ones in the VGG dataset. Besides, the Bray-Curtis distance (BC; \cite{Bray1957-OUF}) was used instead of the Euclidean distance. Since, the BC distance works better for high-dimensional vectors in comparison with the Euclidean and the L1 distances \cite{Sudholt2015-MIA}. The BC distance was measured between the descriptors of two face-images from a set of $3000$ matched and $3000$ non-matched pairs of images. Different from \cite{Parkhi15-DFR} and \cite{Krizhevsky2012-ICD}, the feature vectors from the crops of the image's corners were not utilized for computing the final descriptor, but only the crops from the image's centers. If the BC distance is smaller than a threshold, then the two images portray the same identity. The Equal Error Rate (EER) \cite{Giot2013-FCP,Parkhi15-DFR} was used as the metric, which is defined as the value where the False Acceptance Rate (FAR), and the False Rejection Rate (FRR) are equal.                   
            
Table \ref{tab:maxoutVGG} shows the EER for networks without a maxout layer and with a maxout layer with $k=4$, as well as the results for pruning from one up to three neurons from each maxout unit. 
Similar to the previous results, neurons are pruned, and consequently weights are reduced, from the networks without affecting their performance negatively. 
In fact, the EER decreases by $1\%$ and $0.47\%$ for the VGG16-MFC and the VGG16-MC respectively.
It is assumed that this improvement is produced by the elimination of redundancies in the maxout units. 
Based on a randomization test, the improvement in the VGG16-MFC is highly significant with $p = 0.0026$ \cite{Ojala2010-PTF}. 

The weight reduction changes depending on the position of the maxout layer and on the layer where neurons are pruned in the network. In VGG16-MFC, neurons are pruned from the largest layer in the network $fc6$ reducing the number of weights by $17.7\%$ per each neuron pruning step. Moreover, the maxout layer reduces directly the size of the second largest layer $fc7$. In VGG16-MC on the contrary, neuron pruning does not affect considerably the size of the network, since it reduces a small layer $conv\-5$ compared to $fc6$. The weight reduction comes precisely from the maxout layer, which reduces the size of $fc6$. There is a difference of $7.45\%$ between the weight reduction for both networks, since the size of the layer $fc7$ is never changed in VGG16-MC.

        \begin{table}[t]
            \scriptsize
            \hspace*{-0.7em}
            \centering
            \setlength{\tabcolsep}{2pt}
            \renewcommand{\arraystretch}{1.2}
                \begin{tabular}{c||c||c c||c c||c c||c c}
                    Network & No Maxout & \multicolumn{2}{c||}{No prun} &\multicolumn{2}{c||}{1 neuron prun.} & \multicolumn{2}{c||}{2 neuron prun.} &\multicolumn{2}{c}{3 neuron prun.}\\
                    \hline
                    Type & EER$\%$ & EER$\%$ & p.w.$\%$ & EER$\%$ & p.w.$\%$ & EER$\%$ & p.w.$\%$ & EER$\%$ & p.w.$\%$ \\
                    \hline
                    VGG16-MFC & 3.7 & 3.33 & 8.68 & \textbf{2.83} & 26.39 & \textbf{2.76} & 44.11 & \textbf{2.66} & \textbf{61.82}\\
                    \hline
                    VGG16-MC & 3.7 & 3.7 & 53.15 & 3.36 & 53.56 & 3.36 & 53.96 & \textbf{3.23} & \textbf{54.37}\\
                \end{tabular}
                \vspace*{1em}
            \caption{EER in [$\%$] and pruned weight's proportions for the VGG16 with maxout layer ($k=4$) after the first fully connected layer (VGG16-MFC) and after the last convolutional layer (VGG16-MC).}
            \label{tab:maxoutVGG}
        \end{table}

Additional to neuron pruning, weights from both networks were thresholded after pruning three out of four neurons in the maxout units with the results shown in Table \ref{tab:weightPruning_DVGG}. 
The network's performance will be, deeply, affected if more than $50\%$ for the VGG16-MFC and $30\%$ for the VGG16-MC of the weights are pruned. 
Nevertheless, a total compression rate of $80.1\%$ for VGG16-MFC and $68\%$ for VGG16-MC without performance deterioration can be reached.

    \begin{table}[h]
            \centering
            \setlength{\tabcolsep}{4pt}
            \renewcommand{\arraystretch}{1.1}
            \begin{tabular}{c||c c c c c c c c c}
                Network &\multicolumn{9}{c}{proportions in $[\%]$ of pruned weights}\\
                \hline
                Type & 0 & 10 & 30 & 50 & 60 & 70 & 80 & 90 & 98\\
                \hline
                VGG16-MFC & 2.66 & 2.90 & 3.4 & \textbf{3.34} & 3.5 & 4.0 & 4.53 & 50.00 & 50.00\\
                \hline
                VGG16-MC & 3.26 & 3.77 & \textbf{3.93} & 4.30 & 5.40 & 8.3 & 35.47 & 47.63 & 50.00\\
        \end{tabular}
        \vspace*{1em}
        \caption{EER in [$\%$] for both VGG16 networks, after pruning three out of four neurons, under different proportions of pruned weights.}
        \label{tab:weightPruning_DVGG}
    \end{table}

%% file: conclusion.tex
\section{Conclusion}
We have presented an efficient approach for reducing the size of deep neural networks. 
This approach prunes entire neurons and thus reduces the number of weights in neural networks. 
It uses maxout units for combining $k$ single neurons into complex ones. A maxout layer reduces the number of weights between two adjacent layers by $k$. 
By using these maxout units, the network's performance is not negatively affected, since they boost the dropout benefits reducing redundancies in the network. 
Within these maxout units, neurons are pruned based on a local and non-expensive relevance measure. 
This relevance measure depends on the number of times neurons are maximal for each of the $k$ adjacent input-neurons per maxout unit. 
It differs from previous relevance measures, because it does not depend on the overall network's performance with and without individual neurons \cite{mozer1989-SNN}. 
In general, this approach does not require expensive post-processing, only a single re-training after pruning. 
The performance of this reduction approach depends strongly on the position of the maxout layer in the network. 
As inputs from maxout units are the neurons to be pruned, it is advisable to place the maxout units after the largest layer in the network, because neurons in this layer have large numbers of weights compared with neurons in other layers. 
So, pruning these neurons out of the network is favorable.

By comparing the number of pruned neurons and network's performances between the aforementioned approach and weight pruning, the last one does not delete entire neurons, but rather sets weights to zero. 
Therefore, the architecture's size is only implicitly reduced, and the memory footprint remains equal without a sparse representation.
The proposed approach allows to reduce a network's size by $61 - 74\%$ on an architectural level and without affecting the network's performance.
Assuming a sparse representation, a combination of the proposed neuron pruning with additional weight pruning allows for reducing the size of a network by up to $92\%$.

\section*{Acknowledgment}
This work has been supported by the German Research Foundation (DFG) within project Fi799/9-1 ('Partially Supervised Learning of Models for Visual Scene Recognition').